\documentclass[11pt,a4paper]{article}

\usepackage[dvipsnames]{xcolor}
\definecolor{eamlcolor}{RGB}{0,90,181}
\definecolor{mleacolor}{RGB}{220,50,32}
\usepackage{pifont}
\usepackage{bm}
\definecolor{darkred}{RGB}{204,0,0}

\usepackage{graphicx}
\usepackage[round,sort&compress]{natbib}
\usepackage{hyperref}
\hypersetup{
    colorlinks=true,
    linkcolor=purple,
    filecolor=cyan,      
    urlcolor=teal,
    citecolor=blue,
    }

\usepackage{url}
\usepackage[font=small,labelfont=bf]{caption, subcaption}
\usepackage{multirow}
\usepackage{tabularx}

\usepackage{listings}
\definecolor{codegreen}{rgb}{0,0.6,0}
\definecolor{codegray}{rgb}{0.5,0.5,0.5}
\definecolor{codepurple}{rgb}{0.58,0,0.82}
\definecolor{backcolour}{rgb}{0.98,0.98,0.96}
\lstdefinestyle{mystyle}{
    backgroundcolor=\color{backcolour},   
    commentstyle=\color{codegreen},
    keywordstyle=\color{magenta},
    numberstyle=\tiny\color{codegray},
    stringstyle=\color{codepurple},
    basicstyle=\ttfamily\scriptsize,
    breakatwhitespace=false,         
    breaklines=true,                 
    captionpos=b,                    
    keepspaces=true,                 
    numbers=none,                    
    numbersep=5pt,                  
    showspaces=false,                
    showstringspaces=false,
    showtabs=false,                  
    tabsize=2
}
\title{\bf Task and Explanation Network} 

\author
{Moshe Sipper\\
\\
\normalsize{Department of Computer Science, Ben-Gurion University of the Negev,}\\
\normalsize{Beer-Sheva, 8410501, Israel}\\
\normalsize{www.moshesipper.com, sipper@bgu.ac.il.}
}

\date{\today}

\begin{document} 
\maketitle 

\begin{abstract}
Explainability in deep networks has gained increased importance in recent years. We argue herein that an AI must be tasked not just with a task but also with an explanation of why said task was accomplished as such. We present a basic framework---Task and Explanation Network (TENet)---which fully integrates task completion and its explanation.
We believe that the field of AI as a whole should insist---quite emphatically---on explainability.
\end{abstract}

\section{Explainability in AI}
\label{sec:Explainability}
With the meteoric rise of AI over the past decade, and in particular deep learning, an issue that has been gaining more and more traction is that of \textit{explainability}. An entire subfield of \textit{eXplainable Artificial Intelligence (XAI)} has arisen in recent years \citep{gunning2017explainable}, focusing on finding explanations for the reasoning of deep networks. 

Deep learning pioneer Geoffrey Hinton averred in a recent interview: ``What we did was we designed the learning algorithm. That's a bit like designing the principle of evolution. But when this learning algorithm then interacts with data, it produces complicated neural networks that are good at doing things. But we don't really understand exactly how they do those things.'' \citep{Pelley23}

Since we do not understand how they do those things, we propose the following tenet:
\begin{quote}
    \textit{A deep network should (always) provide an explanation for its output.}
\end{quote}
Towards this end we offer a basic network that fully integrates task completion and explanation.

The next section touches upon the XAI literature.
\autoref{sec:TENet} delineates our Task and Explanation Network---TENet.
\autoref{sec:Results} presents experimental results.
We end with a discussion in \autoref{sec:Discussion}, 
followed by concluding remarks in \autoref{sec:Concluding}.

\section{XAI}
\label{sec:Literature}

It is beyond the scope of this paper to provide a review of XAI literature. Moreover,  several excellent reviews have been offered in recent years \citep{confalonieri2021historical,angelov2021explainable,minh2022explainable,rodis2023multimodal}.

Perhaps the closest to our approach herein is the recent emerging interest in multimodal XAI frameworks. In particular, some frameworks combine images and text, as we do in this paper. Essentially, with multimodel XAI, large subnetworks (or indeed separate networks) attend to the different modalities. 

As recently noted by \citet{rodis2023multimodal} in their survey of multimodal XAI, there are basically three possible processing stages at which explanations are produced:
\begin{enumerate}
    \item Intrinsic: produce explanations by analyzing a primary (task) model's internal structure and parameters.

    \item Post-hoc: approaches that are based on the analysis of the task model’s output.

    \item Separate module: develop an explanation-providing model that is separate from the task model.
\end{enumerate}

As we shall see in the next section our approach is quite different than all three.

\section{TENet: Task and Explanation Network}
\label{sec:TENet}

We describe below the two components of our framework: the dataset and the model.

\paragraph{Dataset.}
We used the Common Objects in Context (COCO) dataset, one of the most popular benchmarking datasets in deep learning \citep{coco}.
It comprises a training set of 118,287 images and a validation set of 5000 images.
There are 91 object classes.
Crucially, for our purpose herein, each image also comes with 5 captions.

The 91 classes underlie the classification task to be performed, which is multi-label---an image may contain more than a single object class.

The captions serve as our explanatory vehicle, providing insight into what the network is ``seeing''. To run TENet we applied the following steps to the captions:
\begin{enumerate}
    \item Generate a raw vocabulary of words from all words in all image captions. This raw vocabulary contains 30,567 words. However, these include 
    14,065 words that appear only once, 19,362 words that appear 3 times or less, 48 words of length 1 character, and 306 words of length 2 characters. 
    
    \item Save only words that appear at least 4 times with length at least 3 characters.

    \item Of these saved words, the final vocabulary is the \texttt{VOCAB\_SIZE} most frequent words, with \texttt{VOCAB\_SIZE} being a hyperparameter (\autoref{tab:hyperparameters}).

    \item Finally, we generate a custom COCO dataset for our purpose, where the ground-truth (multi-label) outputs comprise: 
    i) 91 one-hot-encoded classes, 
    and ii) \texttt{VOCAB\_SIZE} one-hot-encoded word labels.
\end{enumerate}

TENet is expected to classify the objects in an image \textit{and} offer vocabulary words. These latter may be considered as an explanation or at least an explanatory addendum to the multi-label classifier. As recently noted by \citet{prince2023understanding}: ``There is also an ongoing debate about what it means for a system to be explainable, understandable, or interpretable... there is currently no concrete definition of these concepts.''

\paragraph{Model.}
The basic idea behind TENet (Task and Explanation Network) is simple in nature.
We use a common backbone model for both task and explanation, selected from the cornucopia available in \texttt{PyTorch} \citep{Paszke_PyTorch_An_Imperative_2019}.
We only replace the final model layer with \textit{two} heads:
1) a classification head, which predicts object classes, and 
2) an explanation head, which predicts caption words.

\autoref{lst:CombinedModel} shows the code. Essentially, we are enacting what might be referred to as ``weight overloading'', where the network's weights (except for the final layer) must double for both task completion and explanatory rendition.

\begin{lstlisting}[language=Python,upquote=true,caption={TENet model, combing task and explanation.},label=lst:CombinedModel,float=ht]
class CombinedModel(nn.Module):
    def __init__(self, backbone):
        super(CombinedModel, self).__init__()
        self.backbone = backbone

        # Pass a sample through the backbone to get the number of output features
        with torch.no_grad():
            backbone_output = self.backbone(torch.randn(1, 3, HEIGHT, WIDTH))

        # Define separate heads for image classification and caption classification
        self.classification_head = nn.Linear(backbone_output.size(1), NUM_CLASSES)  
        self.caption_head = nn.Linear(backbone_output.size(1), VOCAB_SIZE)  

    def forward(self, images):
        features = self.backbone(images)
        classification_output = self.classification_head(features)
        caption_output = self.caption_head(features)
        return classification_output, caption_output
\end{lstlisting}

Our approach integrates the task and explanation into a single whole through weight overloading. It is thus different than the three approaches to multimodal XAI discussed in \autoref{sec:Literature}.

Once the model is defined as in \autoref{lst:CombinedModel}, standard backpropagation training is applied, with the only difference being the computation of two losses instead of one:
a loss for the output of the classification head and a separate loss for the output of the explanation head. For both losses we used binary cross entropy. 
The sum of the two losses serves as the total loss for the backpropagation phase.

For prediction we applied a ranking-based method, outputting the classes corresponding to the \texttt{TOP\_C} class-output logits, and the words corresponding to the \texttt{TOP\_W} word-output logits. \texttt{TOP\_C} and \texttt{TOP\_W} are hyperparameters (\autoref{tab:hyperparameters}).

The code is available at \url{https://github.com/moshesipper}.

\section{Results}
\label{sec:Results}
We ran over 200 preliminary experiments to ascertain which models and hyperparameters
worked best. An experiment took 1-2 days on our departmental $\sim$100-node cluster (which comprises NVIDIA gtx1080, rtx2080, rtx3090, rtx6000, and rtx4090 nodes).
We experimented with the following 6 backbone models from \texttt{torchvision.models}:
\texttt{ResNet50},
\texttt{MobileNet\_V3\_Small},
\texttt{RegNet\_Y\_400MF},
\texttt{ConvNeXt\_Small},
\texttt{Swin\_V2\_B},
\texttt{and ViT\_B\_16}.

Following the batch of preliminary experiments we decided to focus on the top 2 models that emerged---\texttt{ResNet50} and \texttt{RegNet\_Y\_400MF}---and on the hyperparameters given in \autoref{tab:hyperparameters}.

\begin{table}[ht]
    \centering
    \small
    \begin{tabularx}{0.95\textwidth}{@{}rcX@{}} 
        Name & Value & description \\ \hline
        \texttt{NUM\_CLASSES} & 91   & number of COCO classes (categories)\\
        \texttt{VOCAB\_SIZE}  & 1000 & number of words in vocabulary \\
        \texttt{NUM\_EPOCHS}  & 20   & number of training epochs\\
        \texttt{BATCH\_SIZE}  & 32   & batch size\\
        \texttt{HEIGHT}       & 400  & image height\\
        \texttt{WIDTH}        & 400  & image width\\
        \texttt{TOP\_C}       & 3    & number of top classes to output when making model predictions\\
        \texttt{TOP\_W}       & 10   & number of top explanatory words to output when making model predictions\\
    \end{tabularx}
    \normalsize
    \caption{Hyperparameters.}
    \label{tab:hyperparameters}
\end{table}

There are numerous ways to measure multi-label accuracy, which can be categorized as sample-based, label-based, ranking-based, or hierarchical \citep{Sorower10}. We defined accuracy per predicted image by defining
\textit{task accuracy} and \textit{explanation accuracy}.
Given that our model makes predictions by outputting top-ranked labels, 
task accuracy, $\mathit{acc}_t$,  is defined as:
% sum(1 for i in topc if classes[idx][i] == 1) / TOP_C
\[
\mathit{acc}_t = \frac{1}{\texttt{TOP\_C}} \sum_{i \in \mathit{topc}} y_i,
\]
where 
\texttt{TOP\_C} is the number of top classes to output when predicting (\autoref{tab:hyperparameters}),
$\mathit{topc}$ are the indices of the \texttt{TOP\_C} classes,
and $y_i$ is the ground truth for the $i$th class ($0$ or $1$).
%sum(1 for i in topw if words[idx][i] == 1) / TOP_W
Similarly, explanation accuracy, $\mathit{acc}_e$, is defined as:
\[
\mathit{acc}_e = \frac{1}{\texttt{TOP\_W}} \sum_{i \in \mathit{topw}} y_i,
\]
where 
\texttt{TOP\_W} is the number of top explanatory words to output when making model predictions (\autoref{tab:hyperparameters}),
$\mathit{topw}$ are the indices of the \texttt{TOP\_W} labels (words),
and $y_i$ is the ground truth for the $i$th word ($0$ or $1$).
Overall accuracy per image is simply the arithmetic mean of $\mathit{acc}_t$ and $\mathit{acc}_e$.

\autoref{tab:results} shows the results of 10 runs for each of the two backbone models of focus.
 
\begin{table}[ht]
    \centering
    \small
    \begin{tabular}{r|ccc}
                 & Overall  & Task     & Explanation \\
        Backbone & Accuracy & Accuracy & Accuracy    \\ \hline
        \multirow{10}*{\texttt{ResNet50}} 
			 & 0.582 & 0.643 & 0.521 \\
			 & 0.580 & 0.639 & 0.520 \\
			 & 0.579 & 0.637 & 0.522 \\
			 & 0.579 & 0.636 & 0.522 \\
			 & 0.578 & 0.636 & 0.521 \\
			 & 0.578 & 0.636 & 0.521 \\
			 & 0.577 & 0.637 & 0.518 \\
			 & 0.577 & 0.634 & 0.519 \\
			 & 0.576 & 0.634 & 0.519 \\
			 & 0.575 & 0.631 & 0.520 \\
        
        \hline
        
        \multirow{10}*{\texttt{RegNet\_Y\_400MF}}
			 & 0.591 & 0.642 & 0.539 \\
			 & 0.590 & 0.641 & 0.539 \\
			 & 0.590 & 0.641 & 0.539 \\
			 & 0.590 & 0.641 & 0.539 \\
			 & 0.589 & 0.642 & 0.536 \\
			 & 0.589 & 0.640 & 0.537 \\
			 & 0.589 & 0.639 & 0.539 \\
			 & 0.588 & 0.640 & 0.537 \\
			 & 0.588 & 0.640 & 0.535 \\
			 & 0.587 & 0.638 & 0.536 \\
    \end{tabular}
    \normalsize
    \caption{Accuracy results for 20 runs. Values shown are means over validation-set images.}
    \label{tab:results}
\end{table}

\section{Discussion}
\label{sec:Discussion}
While the accuracy values in \autoref{tab:results} may seem at first blush lower than one might desire, it is important to remember that this is not a single-label scenario. To wit, obtaining even part of the ground-truth labels is useful and informative. 

Consider \autoref{fig:example}, which presents four sample TENet runs over validation-set images. We note that classification outputs are quite coherent, and that explanation words are tenable.
And this, as we saw, is accomplished by a \textit{single, integrated network}.

\begin{figure}[ht]
     \centering
     \begin{subfigure}[b]{0.45\textwidth}
         \centering
         \includegraphics[width=\textwidth]{pic1.jpg}
         \caption{\textbf{Classes}: person, tie, dining table. 
                  \textbf{Words}: man, standing, table, cake, woman, two, wearing, red, front, young.}
         \label{fig:example1}
     \end{subfigure}    
     \hfill
     \begin{subfigure}[b]{0.45\textwidth}
         \centering
         \includegraphics[width=\textwidth]{pic2.jpg}
         \caption{\textbf{Classes}: person, backpack, skis. 
                  \textbf{Words}: people, snow, mountain, man, slope, two, top, person, mountains, men.}
         \label{fig:example2}
     \end{subfigure}     
     \par\bigskip  
     \begin{subfigure}[t]{0.45\textwidth}
         \centering
         \includegraphics[width=\textwidth]{pic3.jpg}
         \caption{\textbf{Classes}: cat, tie, bed. 
                  \textbf{Words}: cat, tie, sitting, top, brown, black, white, striped, bed, close.}
         \label{fig:example3}
     \end{subfigure}     
     \hfill     
     \begin{subfigure}[t]{0.45\textwidth}
         \centering
         \includegraphics[width=\textwidth]{pic4.jpg}
         \caption{\textbf{Classes}: person, tennis racket, chair. 
                  \textbf{Words}: tennis, player, court, racket, woman, playing, racquet, holding, ball, man.}
         \label{fig:example4}
     \end{subfigure}
    \caption{Four input images from the validation set, with the resultant TENet outputs.}
    \label{fig:example}
\end{figure}

As noted in \autoref{sec:TENet}, the nature of what comprises an `explanation' is far from agreed upon. \autoref{fig:example} demonstrates our path herein, which consists of offering added explanatory terms to the network's (task) output.

This is a basic model, with our intent herein being to underscore the importance of coupling tasks with explanations. \autoref{fig:example} demonstrates how weight overloading can be accomplished successfully, rendering our desideratum in \autoref{sec:Explainability}---a network should provide an explanation for its output---realistic.

\section{Concluding Remarks}
\label{sec:Concluding}
We have argued for the full integration of task accomplishment along with an explanation for how the task was accomplished. TENet, presented herein, is a basic reification of said demand.

Future research can proceed along several lines: 
\begin{itemize}
    \item Improve TENet performance.

    \item Test TENet on other datasets.

    \item Tackle other forms of data, including video and audio.

    \item Improve TENet's basic design to offer more substantive explanations.
\end{itemize}

\clearpage
\small
\bibliography{bib}
\bibliographystyle{unsrtnat}

\end{document}